\documentclass[10pt,twocolumn,letterpaper]{article}

\usepackage{cvpr}
\usepackage{times}
\usepackage{epsfig}
\usepackage{graphicx}
\usepackage{amsmath}
\usepackage{amssymb}

\usepackage{epsfig}
\usepackage{amsmath}
\usepackage{amssymb}
\usepackage{microtype}
\usepackage{booktabs} % for professional tables
\usepackage{subfig}
\usepackage{makecell}

\newcommand{\argmax}{\operatornamewithlimits{argmax}}
\usepackage{color}

\usepackage[colorlinks,citecolor={green}]{hyperref}

\cvprfinalcopy % *** Uncomment this line for the final submission

\def\cvprPaperID{****} % *** Enter the CVPR Paper ID here

\begin{document}

%%%%%%%%% TITLE
\title{Revisit Multinomial Logistic Regression in Deep Learning:\\
Data Dependent Model Initialization for Image Recognition}

\author{
    Bowen Cheng$^{1}\thanks{equal contributions}$, \,\,
    Rong Xiao$^{2*}\thanks{This work was done while Rong Xiao, Yandong Guo and Yuxiao Hu were at Microsoft}$, \,\,
    Yandong Guo$^{3}\footnotemark[2]$, \,\,
    Yuxiao Hu$^{3}\footnotemark[2]$, \,\,
    Jianfeng Wang$^{3}$, \,\,
    Lei Zhang$^{3}$ \vspace{3pt}\\
    $^{1}$University of Illinois at Urbana-Champaign, \\
    $^{2}$Ping An Property\&Casualty Insurance Company of China, \\ 
    $^{3}$Microsoft\\
    {\tt\small $^1$bcheng9@illinois.edu}
    \quad \tt\small $^2$xiaorong283@pingan.com.cn \\
    {\tt\small $^3$yandong.guo@live.com, yuxiaohu@msn.com, \{jianfw, leizhang\}@microsoft.com}
}

% \author{First Author\\
% Institution1\\
% Institution1 address\\
% {\tt\small firstauthor@i1.org}
% % For a paper whose authors are all at the same institution,
% % omit the following lines up until the closing ``}''.
% % Additional authors and addresses can be added with ``\and'',
% % just like the second author.
% % To save space, use either the email address or home page, not both
% \and
% Second Author\\
% Institution2\\
% First line of institution2 address\\
% {\tt\small secondauthor@i2.org}
% }

\maketitle
%\thispagestyle{empty}

%%%%%%%%% ABSTRACT
\begin{abstract}
We study in this paper how to initialize the parameters of multinomial logistic regression (a fully connected layer followed with softmax and cross entropy loss), which is widely used in deep neural network (DNN) models for classification problems. As logistic regression is widely known not having a closed-form solution, it is usually randomly initialized, leading to several deficiencies especially in transfer learning where all the layers except for the last task-specific layer are initialized using a pre-trained model. The deficiencies include slow convergence speed, possibility of stuck in local minimum, and the risk of over-fitting. To address those deficiencies, we first study the properties of logistic regression and propose a closed-form approximate solution named regularized Gaussian classifier (RGC). Then we adopt this approximate solution to initialize the task-specific linear layer and demonstrate superior performance over random initialization in terms of both accuracy and convergence speed on various tasks and datasets. For example, for image classification, our approach can reduce the training time by $10$ times and achieve $3.2\%$ gain in accuracy for Flickr-style classification. For object detection, our approach can also be $10$ times faster in training for the same accuracy, or $5\%$ better in terms of mAP for VOC 2007 with slightly longer training.

%Deep neural network (DNN) models usually consist of a highly non-linear feature extractor and a task-specific linear layer, \emph{e.g.} a linear classifier in the context of classification. Recent models use stacks of convolutional layers for feature extractor and multinomial logistic regression (a fully connected layer followed with softmax and cross entropy loss) for linear classifier. With limited training data, most tasks (\emph{e.g.} object detection, semantic segmentation, fine-grained classification) require fine-tuning from a network pretrained on a larger dataset by copying parameters of feature extractor and re-initialize parameter of the task-specific linear layer. However, due to the lack of closed-form solution, the task-specific linear layer is usually randomly initialized and finetuned at a lower learning rate, leading to several deficiencies including slow convergence speed, possibility of a stuck in local minimum and the risk of over-fitting. To address this problem, we first study the properties of logistic regression and come up with a closed-form approximate solution named regularized Gaussian classifier (RGC). Then we adopt this approximate solution to initialize the task-specific linear layer and demonstrate superior performance over random initialization in terms of both accuracy and convergence speed on various of tasks and datasets.
\end{abstract}

%%%%%%%%% BODY TEXT
\section{Introduction}

Training a deep neural network is generally solving a non-convex optimization problem
over millions of parameters with no analytical solutions. When there are large scale training data, e.g. ImageNet \cite{ILSVRC15} or millions of face images \cite{guo2016msceleb},
typically people train a DNN model from scratch with an iterative solver, 
which takes days, or up to weeks even with the great progresses 
in computing infrastructure and optimization method. On the other hand, 
when training a DNN model for a specific task (\emph{e.g.} object detection, semantic segmentation, fine-grained classification) which tends to have smaller scale of training data,
fine-tuning (sometimes called transfer learning) is widely adopted. 

In this paper, we focus on improving the fine-tuning method in terms of both accuracy and training speed. 
Conventionally, in the fine-tuning schema, the parameters of the lower level layers of the model to be trained 
are transfered directly from a pre-trained model, \emph{e.g.} AlexNet \cite{AlexHinto_DNN}, ResNet \cite{MSRA_150}, which is trained on a much larger scale dataset, \emph{e.g.} ImageNet-1k \cite{ILSVRC15} classification dataset, while the parameters of the last layer 
are randomly sampled from certain distributions (usually Gaussian) \cite{randominit} and are optimized together with previous layers in a multinomial logistic regression manner (\emph{i.e.} a fully connected layer followed with softmax and cross entropy loss). 
Examples include Flickr style estimation \cite{Flickr}, flower recognition \cite{Flower}, and places recognition \cite{Places}. Fine-tuning schema can also be used in other image recognition domain. For example, image classification models trained on ImageNet can also be used to initialize an object detection model, such as YOLO \cite{yolo9000} and Faster R-CNN \cite{frcnn}, and lead to better performance.

The conventional fine-tuning method successfully leverages the 
low level visual pattern extractors learned from general tasks (classification), 
which reduces the over-fitting issue to some extend, 
and typically converges faster than training from scratch. 
However, the conventional fine-tuning still suffers from the following two challenges
because new layers are randomly initialized. 
The first is still over-fitting. 
%If all the millions of parameters for the network to be fine-tuned are free to be updated, 
%the feasible region is usually too large for a specific task with limited amount of training data. 
Even locking the lower level layers and only updating new layers
still has the over-fitting issue. The reason might be that 
the dimension of the input (called features) to the last layer 
%input feature dimension 
is generally too high compared with the limited training data, and setting the parameters of 
new layers randomly at the initial stage might be too far away from the optimal solution.    
For example, the input of the 'fc8' layer in VGG \cite{vgg_DNN} or AlexNet \cite{AlexHinto_DNN} is a tensor with 4096 channels and the input of 'fc' layer in ResNet-50 \cite{MSRA_150} is a tensor with 2048 channels. 
Our experiments with various tasks validate this, as discussed in the experimental results section. 

The second challenge of the conventional fine-tuning is the convergence speed.
Though for image classification, fine-tuning can converge very fast, for more complex tasks like object detection, it still needs hours or days to fine-tune from pretrained models \cite{frcnn,yolo9000}. This is mainly because
with a randomly initialized linear layer, 
one has to use a smaller learning rate for the parameters in the non-linear feature extraction layers to avoid gradients of randomly initialized parameters in new layers ruining the pre-trained model. 
This is impractical or inefficient for applications that require frequent model training or prototyping. 
For example, for a web-based training service like Microsoft Custom Vision\footnote{http://customvision.ai}, 
the model needs to be trained within couple of minutes to guarantee the user experience and the productivity.

The above problems are mainly caused by the fact that the newly added layer is randomly initialized due to the lack of closed-form solution of logistic regression. Recent research \cite{goodfellow6572explaining} has shown that DNN models have near linear decision boundaries. Thus we can assume the pretrained feature extractor is general enough to obtain near linearly separable features on the new training set. As a result, we can leverage solutions from other linear classifiers, e.g. linear discriminative analysis (LDA), to initialize the linear layer in logistic regression, with the hope that a well-initialized linear layer is closer to the optimum than a randomly initialized linear layer. We have explored several linear classifiers and found all of them give reasonably good results. Among these classifiers, logistic regression gives the best result which makes sense because logistic regression is more consistent with the loss function (cross entropy loss) in DNN training. However, the problem is that logistic regression is time-consuming since it does not have closed-form solution and needs an iterative solver.

In order to tackle the above drawbacks of logistic regression, we first study the properties of logistic regression and find that it has an exponential family distribution. We further show that the exponent is an infinite order polynomial with zero and first order coefficients class-dependent and all other higher order coefficients class-independent. 
Based on this finding, we can approximate the distribution of the features 
for each class with a special case of exponential family with polynomial exponent, \emph{i.e.} Gaussian distributions. To satisfy the coefficient constraints, the Gaussian distribution should have a class-dependent mean vector (first-order coefficients) and a class-independent covariance matrix (second-order coefficients). 
We derive \textbf{a closed-form solution} of the optimal linear classifier by maximum likelihood (ML) under this approximation named \emph{regularized Gaussian classifier} (RGC). RGC can be served as an approximate solution to multinomial logistic regression in DNNs while having the advantages that it is fast, it has a closed-form solution, and it is hyper-parameter free.
% \jianfw{while -> but it is fast and free of hyper-parameters with a closed form solution. }
This linear classifier is then used to initialize the last layer of the DNN model. That is, we copy the solution of RGC ($\mathbf{w}_k$ and $b_k$ for each of the $K$ categories, $k=1,\dots,K$) to initialize the parameters in the last linear layer in the DNN model.
%a regularized Gaussian classifier (RGC) to do data-dependent model initialization. 
Compared with the random initialization algorithm, the proposed algorithm can dramatically reduce the training cost and lead to a better model with negligible initialization cost. 

We have applied the proposed model initialization algorithm to both the image classification and object detection problems. Extensive experiment results  demonstrate the superiority of the method in both scenarios. 
For example, for image classification, our approach can reduce the training time by $10$ times and achieve $3.2\%$ gain in accuracy for Flickr-style classification. For object detection, our approach can also be $10$ times faster in training for the same accuracy, or $5\%$ better in terms of mAP for VOC 2007 with slightly longer training.

\section{Related Works}
\label{sec:research_background}

\subsection{Model Fine-tuning}
A typical deep neural network model can be decoupled into two parts, 
a non-linear feature extraction part $\phi(\cdot)$ which corresponds to a stack of layers, 
followed by a linear classification part. %, as shown in Fig. \ref{fig:dnns}.
The assumption that the feature extraction part can extract some general task-independent features gives rise to the possibility of fine-tuning. 
Existing fine-tuning schema mainly fall into two categories: 1) randomly initializing the last linear layer and fine-tuning both the linear and non-linear stages (entire model) on the new data set \cite{jia2014caffe,Flower,Places}, 
and 2) fixing the model parameters in the feature extraction stage and training a linear classifier, such as linear SVM, or multinomial logistic regression, for the new task \cite{Yoshua:nips2014}. 
The second category could be very fast for some lightweight image classification tasks, 
but suffers from suboptimal accuracy on test sets. 
%It is well known that learning a linear classifier in high dimensional space is very challenging.
As shown in \cite{Yoshua:nips2014}, the authors 
%first train a network with one set of training data, 
%then
%artificially remove the last layer (linear classification stage) with the parameters 
%from the non-linear stage fixed. 
retrained the last layer (linear classification stage) with the parameters 
from the non-linear stage fixed and obtained suboptimal accuracy. 
%and retrain the linear classification layer
%using the same training data. 
%fix the model parameters in the feature extraction layers and re-train the last logistic regression layer on the same data set, 
%Experimental results show that it leads to suboptimal results. 
In order to overcome the problem of over-fitting, various model regularization methods, for example, weight decay, have been proposed. However, weight decay degrades the efficiency of SGD optimization and may lead to the under-fitting problem. In this paper, we find that our proposed model initialization method prevents models from over-fitting during fine-tuning, without tuning any hyper parameters.

\subsection{Model Initialization}
Previous works initialize models from a predefined distribution (also known as random initialization), \emph{e.g.} uniform distribution or Gaussian distribution with different statistics (\emph{i.e.} mean and standard distribution of a Gaussian distribution) to deal with the vanishing gradient problem \cite{randominit,Glorot10understandingthe,he2015}. More recent works have placed attentions on data-dependent model initialization. \cite{krahenbuhl2015data} proposed to normalize randomly initialized weights according to the training data in order to let parameters to learn at the same rate (activations are equally distributed), finally, they resacle each layer such that the gradient ratio is constant across layers. However, the initialization is still stochastic in \cite{krahenbuhl2015data}. In this paper, we propose a deterministic data-dependent model initialization method and we find our method increase the convergence speed significantly. \cite{seuret2017pca} proposes a PCA-based model initialization method where PCA is first turned into an auto-encoder, then the auto-encoder is trained and the weights are used to initialize the model. However, there is no closed-form solution for the auto-encoder and the training requires tuning many hyper-parameters. Compared with this method, our proposed method has a closed-form solution thus it does not require tuning any hyper-parameter. 

% \leizhangc{This section looks too simple. The last two sentences starting from "In order to overcome" do not seem bringing any useful information to our work. In addition, we'd better to discuss the "data-dependent model initialization" paper as pointed by reviewers.}

\section{Logistic Regression in Deep Learning}
\label{sec:logistic}

\subsection{Multinomial Logistic Regression Revisit}
Softmax with cross-entropy loss is widely used in modern DNN network structures. It is well-known that a single fully connected neural network with Softmax and cross-entropy loss is equivalent to multinomial Logistic regression. Suppose that $\mathbf{x}$ is the input of the network, $k\in K$ is the class label, $\mathbf{w}_k$ and $b_k$ are network parameters associated with class $k$. Then the probability of $\mathbf{x}$ belonging to the class $k$ can be defined by the Softmax function:
\begin{equation}
p(k|\mathbf{x}) =\frac{1}{z(\mathbf{x})}\exp(\mathbf{w}_k^T\mathbf{x}+b_k) \,,
\label{eq:softmax}
\end{equation}
where $z(\mathbf{x}) = \sum_{j=1}^{K}\exp(\mathbf{w}_j^T\mathbf{x}+b_j)$.
Then the likelihood function of class $k$ will be:
\begin{equation}
p(\mathbf{x}|k) = \exp(\mathbf{w}_k^T\mathbf{x} + b_k' - \epsilon(\mathbf{x})) \,,
\label{eq:classprob}
\end{equation}
where terms $\mathbf{w}_k$ and $b_k'= b_k-\log(p(k))$, which only depend on class $k$, and $\epsilon(\mathbf{x})=\log(z(\mathbf{x}))-\log(p(\mathbf{x}))$, which only depends on $x$. Let $\mathcal{F}(\mathbf{x}, k) = \mathbf{w}_k^T\mathbf{x} + b_k'$, Eq. \ref{eq:classprob} can be written as
\begin{equation}
p(\mathbf{x}|k) = \exp(\mathcal{F}(\mathbf{x}, k) - \epsilon(\mathbf{x}))
\label{eq:classprob_final}
\end{equation}
Let the log-likelihood function be $q(\mathbf{x}, k) = \log(p(\mathbf{x}|k)) = \mathcal{F}(\mathbf{x}, k) - \epsilon(\mathbf{x})$. We notice that $q(\mathbf{x}, k)$ has a class-dependent part $\mathcal{F}(\mathbf{x}, k)$ which is linearly dependent of class $k$ and a class-independent part $\epsilon(\mathbf{x})$.
If we use multi-variable Taylor series to represent $q(\mathbf{x}, k)$ at point $\mathbf{a}$, we will have:
\begin{equation}
\begin{aligned}
q(\mathbf{x}, k) &= q(\mathbf{a}, k) + \nabla q(\mathbf{a}, k)(\mathbf{x}-\mathbf{a})\\ &\quad + \frac{1}{2}(\mathbf{x}-\mathbf{a})^THq(\mathbf{a}, k)(\mathbf{x}-\mathbf{a}) + ...
\label{eq:talor}
\end{aligned}
\end{equation}
% \begin{equation}
% q(\mathbf{x}, k) = q(\mathbf{a}, k) + \nabla q(\mathbf{a}, k)(\mathbf{x}-\mathbf{a}) + \frac{1}{2}(\mathbf{x}-\mathbf{a})^THq(\mathbf{a})(\mathbf{x}-\mathbf{a}) + ...
% \label{eq:talor}
% \end{equation}
Since the class-dependent term is only linearly dependent on class $k$, meaning the second and higher order derivatives in Eq. \ref{eq:talor} are independent of class $k$. From Eq. \ref{eq:classprob_final} and \ref{eq:talor}, we can conclude that logistic regression has the following properties:
\begin{enumerate}
\item The examples in each class follow a exponential family distribution.
\item The exponent is a polynomial with infinite order and only the coefficients of zero and first order are dependent on class.
\end{enumerate}

\subsection{Gaussian Classifier as Logistic Regression}

It is well known that there is no analytical solution for logistic regression. However, if we ignore the 3rd and higher order terms, the samples in each class can be assumed to follow a Gaussian distribution with some specific requirements. Since the second order term in a Gaussian distribution only depends on its covariance matrix, a class-independent second order coefficient indicates a class-independent covariance matrix.

If we assume that the features from the same class for the last linear layer come from a Gaussian distribution with a class-specific mean vector and a shared covariance matrix $\Sigma$, then the resulting Gaussian classifier becomes logistic regression. Based on this assumption, let $\{\mathbf{x}_i,y_i\},i=1,2,...,N$, $y_i\in K$ denote the features and class labels for the last linear layer. The class centroids $\mu_k$ can be computed as $\mu_k=\frac{1}{|C_k|}\sum_{i\in C_k}\mathbf{x}_i$, where $C_k$ is the set of indices of samples belonging to class $k$. The likelihood function can be evaluated by 
\begin{equation}
P(\mathbf{x}|k)=|2\pi\Sigma|^{-\frac{1}{2}}\exp{\left(\frac{-(\mathbf{x}-\mu_k)^T\Sigma^{-1}(\mathbf{x}-\mu_k)}{2}\right)}
\label{gaussian_likelihood}
\end{equation}

Although it is only an approximation of logistic regression, Gaussian classifier has the advantage that it has a closed-form solution. Even if the solution is not optimal, it is reasonable to use this solution to initialize the task-specific last layer in fine-tuning because the additional optimization steps can push this sub-optimal solution towards the optimal solution.

Our experimental results show that even without fine-tuning, a Gaussian classifier based model initialization can already achieve a promising accuracy. Moreover, as the DNN model is fully initialized, the model parameters can be fine-tuned together with the same learning rate, leading to a even faster convergence speed.

\section{Model Initialization with RGC}
\label{sec:RGC}
\subsection{Regularized Gaussian Classifier (RGC)}
Suppose classes are uniformly distributed in real world, then maximum a posterior (MAP) classifier is the same as maximum likelihood (ML) classifier due to the fact that:
$$p(k|\mathbf{x}) = \frac{p(\mathbf{x}|k)p(k)}{p(\mathbf{x})} \propto p(\mathbf{x}|k)$$ 
Based on the assumption the second order coefficient in the exponent is class-independent, we can assume features are from a Gaussian distribution with class dependent mean and class independent covariance matrix. Thus, we assign a class label to the sample by maximizing the likelihood function of the underlying Gaussian distribution in Eq. \ref{gaussian_likelihood}:
\begin{equation}
\hat{y} = \argmax_{k\in K}p(\mathbf{x}|k) 
\label{eq:ncc}
\end{equation}

Since the quadratic term $x^Tx$ and the exponential function does not affect the order, we can rewrite Eq. \ref{eq:ncc} in the linear form:
\begin{equation}
\hat{y}=\argmax_k\mu_k^T\Sigma^{-1}\mathbf{x}-\frac{1}{2}\mu_k^T\Sigma^{-1}\mu_k
\label{eq:ncc-linear}
\end{equation}

Let
\begin{equation}
\mathbf{w}_k=\Sigma^{-1}\mu_k , \qquad
b_k=-\frac{1}{2}\mathbf{w}_k^T\mu_k
\label{eq:ncc_ws}
\end{equation}

Eq. \ref{eq:ncc-linear} becomes 
\begin{equation}
\hat{y}=\argmax_{k\in Y}\mathbf{w}_k^T\mathbf{x}+b_k
\label{eq:ncc_final}
\end{equation}

Therefore, Eq. \ref{eq:ncc_final} shows an optimal solution to a linear classifier for the problem. However, many real applications do not have sufficient training data. The covariance matrix estimation becomes highly variable, and the weights estimated by Eq. \ref{eq:ncc_ws} is heavily weighted by the smallest eigenvalues and their associated eigenvectors. In order to avoid this problem, we introduce a regularization term to the covariance matrix. We have $\mathbf{w}_k=(\Sigma+\epsilon I)^{-1}\mu_k$, where $I$ is an identity matrix and $\epsilon$ is the regularization term (a pre-chosen small constant, we set it to 0.1 for most of our experiments).  In practice, $\mathbf{w}_k$ can be efficiently calculated by solving the following equation,
\begin{equation}
(\Sigma+\epsilon I) \mathbf{w}_k=\mu_k \, .
\label{eq:solve_w}
\end{equation}

Using Eq. \ref{eq:solve_w}, we avoid the calculation of matrix inverse, which is usually 2-3 times faster in practice.

\subsection{Parameter Calibration}
Generally, for any constant $\alpha>0,\beta$ and constant vector $v$, we can define an infinite set
of weights and biases $\{\hat{\mathbf{w}_k},\hat{b_k}\}$, where
\begin{equation}
\hat{\mathbf{w}_k}=\alpha \mathbf{w}_k + v ,\qquad
\hat{b_k}=\alpha b_k + \beta
\label{eq:ncc_ws2}
\end{equation}
We can prove that all these parameters are equivalent in terms of classification accuracy. However, their impact on SGD optimization will be different. Note that multi-class logistic regression is normally implemented as fully connected layer followed by Softmax with cross entropy loss layer in most deep learning platforms \cite{jia2014caffe}. If we increase $\alpha$ by ten times, the cross-entropy loss after the Softmax operation will be changed, and the loss propagated to previous layers will be changed as well. However, there is no analytical solution to finding an optimal set of parameters which can minimize the cross entropy loss. Instead of solving it directly, we use the weights $\{\mathbf{w}_k'\}$ of the last linear layer in the pre-trained network as the reference. We want both networks to have similar scales of loss which can be properly propagated to previous layers. Therefore, we align $\hat{\mathbf{w}}\in \{\hat{\mathbf{w}_k}\}$ and $\hat{b}\in \{\hat{b_k}\}$ to $\mathbf{w}'\in \{\mathbf{w}_k'\}$ and  $b'\in \{b_k'\}$ as follows.
\begin{eqnarray}
\label{eq:ncc_wavg1}
E(\hat{\mathbf{w}})&=&E(\mathbf{w}')\,,	\\
\label{eq:ncc_wavg2}
E(\hat{b})&=&E(b')\,,	\\
\label{eq:ncc_wavg3}
E(\|\hat{\mathbf{w}}-E(\hat{\mathbf{w}})\|^2)&=&E(\|\mathbf{w}'-E(\mathbf{w}')\|^2)\,,
\end{eqnarray}
where $E(.)$ denotes expectation. From Eq. \ref{eq:ncc_ws2} and Eq. \ref{eq:ncc_wavg1}-\ref{eq:ncc_wavg3}, we have
\begin{eqnarray}
\label{eq:ncc_regv}
v & = & E(\mathbf{w}')-\alpha E(\mathbf{w}) \,, \\
\label{eq:ncc_regb}
\beta & = & E(b')-\alpha E(b)\,, \\
\label{eq:ncc_rega}
\alpha & = & \sqrt{\frac{ E(\|\mathbf{w}\|^2)-\|E(\mathbf{w})\|^2}{ E(\|\mathbf{w}'\|^2)-\|E(\mathbf{w}')\|^2}}\,.
\end{eqnarray}

Based on Eq. \ref{eq:ncc_ws2}, \ref{eq:ncc_regv}-\ref{eq:ncc_rega}, we can get the optimal set of weights $\{\hat{\mathbf{w}_k}, \hat{b_k}\}$ to initialize the last FC layer of a DNN model.

\subsection{Relationship to Other Data Based Methods}

We can find that nearest centroid classifier (NCC) \cite{tibshirani2002diagnosis} is a special case of RGC when $\epsilon\to\infty$. When $\epsilon\to 0$, RGC is very similar to LDA for binary classification. Compared with the multi-class LDA algorithm, RGC omits the calculation of between-class scatter matrix. Using this trick, RGC avoids two times of SVD calculation, which makes RGC 5-20 times faster in practice than the traditional LDA algorithm. The weakness of RGC is that it is incapable of performing dimension reduction, which is not an issue for the model initialization problem.
The regularization used in RGC is not something entirely new. It have been frequently used in different situations, for example, regularized LDA \cite{rlda}, ridge regression, etc.

There are also many other linear classification algorithms, such as support vector machine and Gaussian Process. Compared with logistic regression with SGD, most of these algorithms are slow, especially in a high dimensional feature space with non-linear separable data.

\section{Experimental Results}
\label{sec:exp}

In this section, we evaluate our method on several tasks including finetuning, \emph{e.g.} fine-grained recognition and object detection. We compare our method with other data-independent and data-dependent methods thoroughly on these tasks. Experimental results demonstrate the effectiveness of our method. In the following experiments, all speeds are tested on a DGX-1 sever using one NVIDIA Tesla P100 GPU and Intel E5-2698v3 CPU.

\subsection{Convergence of class covariance matrices}

To demonstrate the covariance matrix of approximated Gaussian classifier is independent on class, we train a ResNet-18 network on the ImageNet dataset for $650k$ iterations. We use SGD with a mini-batch size of $256$. The learning rate starts from $0.1$ and divided by $10$ for each $150k$ iterations. For every $20k$ iterations, we extract DNN features from the ``Pool5'' layer for every images in the ImageNet test set. Then we use the correlation matrix distance (CMD) \cite{herdin2005correlation} metric, which is defined in Eq. \ref{eq:cmd}, to measure the similarity of the covariance matrices between different image categories. 
\begin{equation}
d_{coor}(R_1,R_2)=1-\frac{tr\{R_1R_2\}}{\|R_1\|_f\|R_2\|_f} \, ,
\label{eq:cmd}
\end{equation}
where $R_1$ and $R_2$ are two covariance matrices, $tr\{.\}$ is the trace of a metric, $\|.\|_f$ is the Frobenius norm of a matrix.

\begin{figure}[t]
	\centering
    \includegraphics[width=1\linewidth]{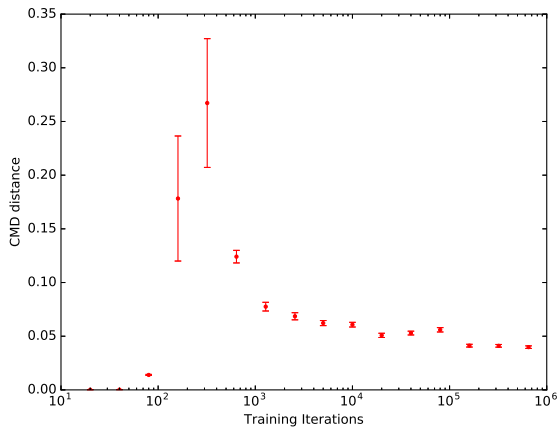} 
	\caption{The study of convergence of class covariance matrices over different iterations. }
	\label{fig:cmdstudy}
\end{figure}

Since there are only $50$ images for each category in the ImageNet validation set, dimension reduction is required to get robust estimation of the covariance matrices in the feature space. Thus we (i) pre-process the feature vector by subtracting its corresponding class mean vector to improve the efficiency of the dimension reduction; (ii) use PCA to reduce the dimensionality of feature vector from $512$ to $2$, (iii) calculate the covariance matrices for each categories and calculate the mean of these covariance matrices; (iv) for each category, calculate the CMD distance between the class covariance matrix and the mean covariance matrix; and (v) calculate the mean and the variance of these CMD scores. The final results are shown in Fig. \ref{fig:cmdstudy}.

From Fig. \ref{fig:cmdstudy}, we find that in the initial training stages, the covariance matrices from different categories are quite similar. The main reason is due to the model random initialization method which is homogeneous. As the training continues, the model starts to fit to the training data, and the mean of the CMD score reaches its maximum at the 320th iteration. After that, the mean and variance of the CMD scores continually drop until converging to the value around 0.04. This result verifies our conclusion that CNN is capable of learning feature to meet the requirement of logistic regression, and as a result the covariance matrices tend to be class-independent under Gaussian approximation. However, it does not converge exactly to zero because the distribution is not exactly Gaussian.

\subsection{Fine-grained Classification}

We choose the Flickr-style dataset \cite{Flickr} for ablation study to evaluate our method on the fine-grained classification task due to its popularity in transfer learning. Flickr-style dataset contains $80,000$ images labeled with $20$ different visual styles. We use the same setting as \cite{jia2014caffe}: 80\% for training and 20\% for testing and fine-tune a AlexNet \cite{AlexHinto_DNN} for fine-grain classification which is pre-trained on ImageNet \cite{ILSVRC15} for 100,000 iterations and achieved the final validation accuracy of 39.16\%. It takes ~7 hours using Caffe on a K40 GPU. 

There are two fine-tuning strategies, one is only fine-tuning the last FC layer while fixing all the previous layers, and the other is fine-tuning both the pretrained layers and the last FC layer together. The first method is usually used when there is not enough data, it treats pretrained layers as feature extractor and train a separate task-specific linear layer, \emph{e.g.} an SVM, on the new dataset. It is fast but usually suffers from over-fitting. The latter treats pre-trained layers as model initialization and optimize both feature extractor and linear classifier simultaneously but with different rate, the latter is slow but usually generalizes better. If otherwise stated, experiments are done with the second setting.

\subsubsection{Data-dependent model initialization details}

In this section, we discuss the implementation details about data-dependent model initialization methods. First, we drop the last task-specific layer of AlexNet and feed the entire training set to the pre-trained Alexnet to extract features from the output of the ``fc7'' layer. Then, we use these extracted features to initialize the newly added task-specific layers following Eq. \ref{eq:ncc_regv}-\ref{eq:ncc_rega}.

\subsubsection{Comparison with other initialization methods}

We compare our proposed RGC with different initialization methods, including RAND (the MSRA random initialization method \cite{he2015}), LDA (linear discriminative analysis), LR (multinomial logistic regression), SVM (linear support vector machine classifier). In this experiment, implementations of LDA, LR and SVM are from the scikit-learn Python package \footnote{http://scikit-learn.org}. In Tab. \ref{tab:initcmp}, we evaluate the accuracy after model initialization \textbf{without fine-tuning}. Comparing RGC with other model initialization methods, RGC is extremely fast, which only takes 3.34 seconds not considering feature extraction time. Another advantage of RGC algorithm is that \textbf{RGC is hyper-parameter free and insensitive to regularization}. Compared with LR and SVM, the RGC algorithm is almost parameter free (we use a fixed $\epsilon=0.1$ for most scenarios), which is critical for online services when a customer lacks machine learning experience. On the other hand, \textbf{LR and SVM algorithms require a lot of effort to manually select dataset-dependent hyper-parameters} (\emph{e.g.} learning rate) to get optimal results.

\begin{table}
\caption{Model initialization accuracies (without fine-tuning) and corresponding costs (not considering feature extraction time). RGC: regularized Gaussian classifier (ours), SVM: support vector machine, LDA: linear discriminative analysis, LR: multinomial logistic regression.}
\centering
\begin{tabular}{l|c|c|c}
Algorithm   & Accuracy & Time (s) & \makecell{Hyper-Parameter\\Selection}\\
\hline
RGC            & 38.11 & \textbf{3.34} & No\\
SVM     	   & 37.27 & 13.78 & Yes\\
LDA			   & 37.42 & 24.73 & No\\
LR    		   & \textbf{38.57} & 97.02 & Yes\\
\end{tabular}
\label{tab:initcmp}
\end{table}

\begin{table}
\centering
\caption {RGC model initialization for Flickr-style }
\begin{tabular}{l|c|c}
Algorithm  &  Iter &  Acc (\%)   \\
\hline
Baseline \cite{jia2014caffe} & 100000 & 39.16  \\
RGC		&	0 &  37.96\\
RGC		&	3000 &  39.20\\
RGC		& 10000 & \textbf{42.39} \\
\end{tabular}
\label {tb:flickre2e}
\end{table}

Next, we fine-tune the models after initialization. We conduct experiments with RGC model initialization described in Eq. \ref{eq:ncc_regv}-\ref{eq:ncc_rega} and let all parameters to update with the same learning rate.
% \textbf{Fine-tune the whole network.} In the second setup, we allow that the parameters from whole network can be updated. 
This setup makes a fair and closer comparison with \cite{jia2014caffe}. More specifically, we follow all settings from \cite {jia2014caffe}, except for the following two parameters. First, since our model is properly initialized, we do NOT use $10$ times of learning rate to update the parameters in the last linear layer. Second, since the initial model is already close to the optimal solution, we reduce the step size to $3000$. The experimental result is shown in Tab. \ref{tb:flickre2e}. With the help from the RGC model initialization method, the proposed approach achieves $37.96\%$ top-1 accuracy in the initial state, and $39.20\%$ at $3000$th-iteration. Finally, the model achieves $42.39\%$ top-1 accuracy, which is significantly higher than the best published results with the same network structure. 

We hypothesize that there are two reasons that might contribute to the gain:
\begin{enumerate}
\item 
RGC-initialized weights are more consistent in distribution as pre-trained weights (this is because they have higher accuracy after initialization), which makes all parameters learning at the same rate during fine-tuning. As discovered in \cite{krahenbuhl2015data}, it is important to have all parameters in the network to learn at the same rate. Although the learning rate of RAND-initialized layers are scaled by 10, they are not guaranteed to learn at the same rate.
\item
We find that RGC-initialized models are less likely to over-fit and we argue that some of the accuracy gap is due to over-fitting. Details about why RGC-initialization reduces over-fitting are discussed in Appendix.
\end{enumerate}

\subsubsection{More results}
We also evaluate our RGC model initialization method on more fine-grained classification datasets (Flower-102 \cite{Flower} and Caltech-256 \cite {griffinHolubPerona}). Compared with state-of-the-art results, using the same pre-trained deep learning model, our model initialization method shows 2-4 times faster in convergence iterations with better results, as shown in Tab. \ref{tb:more1} and  \ref{tb:more2} 

\begin{table}[h]
\centering
\caption {RGC model initialization for Oxford flower-102 }
\begin{tabular}{l|c|c}
Algorithm  &  Iter &  Acc (\%)   \\
\hline
Baseline \cite{caffe-oxford102}  & 16000 & 93.04  \\
RGC		&	0 &  81.37 \\
RGC		& 9000 & \textbf{93.33} \\
\end{tabular}
\label {tb:more1}
\end{table}

\begin{table}[h]
\centering
\caption {RGC model initialization for Caltech-256 data set }
\begin{tabular}{l|c|c}
Algorithm  &  Iter &  Acc (\%)   \\
\hline
Baseline \cite{zeiler2014visualizing}  & 7100 & 73.50 \\
RGC  & 0 &  63.86 \\
RGC & 1600 & \textbf{73.91} \\
\end{tabular}
\label {tb:more2}
\end{table}

% \subsection { Multi-label Binary Classification}
% \begin{figure}[ht]
% 	\centering
% 	\subfloat[Testing Accuracy]{\includegraphics[width=0.9\linewidth]{figs/voc20_accuracy.eps}} 
%     % \subfloat[Training Loss]{\includegraphics[width=0.9\linewidth]{figs/voc20_loss.eps}}	
% 	\caption{ Multi-label binary classification on VOC2007 data set.}
%     %Test accuracy and training loss for the multi-label binary classification task with the VOC 2007 data. Different initial conditions were obtained with different methods. 
% 	\label{fig:classb}
% \end{figure}

% We test our method in the multi-label classification scenario. We use the PASCAL VOC 2007 dataset \cite{pascal-voc-2007}, which includes 5011 training images and 4952 test images containing objects from 20 categories. By removing the bounding box and keep the image level label, we construct a multi-label classification data set. In this experiment, we use the same AlexNet \cite{AlexHinto_DNN} trained with ImageNet \cite{ILSVRC15}. We compared four model initialization technologies,  RGC, NCC, RAND and RGC-ML which are described in Sec. \ref{sec:ml}. After the model initialization,  we use the setting of (weight decay:0.5, batch size:256, learning rate: 0.0001) to fine-tune the model for 100 iterations.

% As shown in  \ref{fig:classb}, the RGC-ML model slightly outperforms the RGC model. It achieves the initial classification accuracy of $73.18\%$ which is close the best baseline accuracy of $75.47\%$ after 100 iterations. The final accuracy of RGC-ML achieves $76.8\%$, which outperforms the baseline algorithm by 1.3\%.

\subsection{Object Detection}
Recent object detectors can be categorized into two types: two-stage detectors, e.g. Faster R-CNN \cite{frcnn} and its variants \cite{dai2016r,lin2017feature,he2017mask,cheng2018revisiting,wei2018ts2c}, which generates region proposals first and then focuses on the recognition of the proposals; and one-stage detectors, e.g. YOLO \cite{redmon2016you,yolo9000} and SSD \cite{liu2016ssd}, which directly predicts the bounding box coordinates and the classification scores. However, what they have in common is that both types of detectors have separate classifiers to classify regions and bounding box regressors to predict object locations.

In the following experiments, we demonstrate that our method works well for both the two-stage (Faster RCNN) and one-stage (YOLO V2) detectors.

\subsubsection{Implementation details}
To demonstrate the effect of model initialization for the classification layer, we modify detectors to have class agnostic bounding box regression so that the bounding box regression layer does not depend on object class. Then we use the pre-trained detector (on ImageNet 200 detection set) to initialize bounding box regression layer in all experiments. For our baseline models, we follow their original implementation to initialize the newly added classification layers randomly with Gaussian distribution. We use the RGC initialization to initialize the classification layer as our proposed method. In the following experiments, we demonstrate the effectiveness of our method on YOLO V2 with Darknet-19 backbone \cite{yolo9000}, and Faster R-CNN with ZF backbone \cite{ren2015faster}.

\begin{figure}[h]
	\centering
	\subfloat[YOLOV2]{\includegraphics[width=0.9\linewidth]{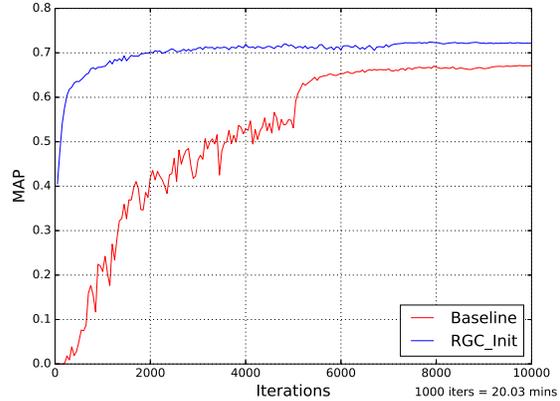}}  \\
	\subfloat[Faster R-CNN]{\includegraphics[width=0.9\linewidth]{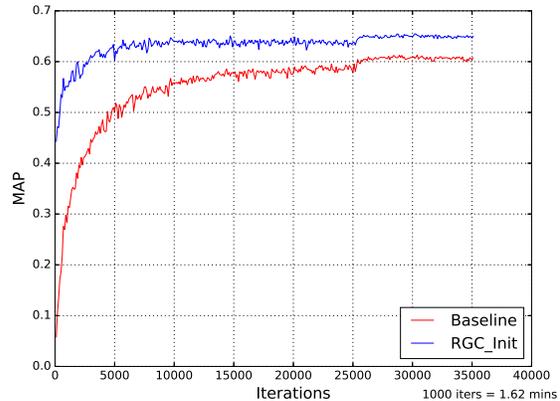}}
	\caption{Model initialization for (a) YOLO V2 and (b) Faster R-CNN on PASCAL VOC 2007 dataset}
	%\leizhangc{in (b), the two curves end at iter 35000. why not to let the x-axis end at iter 35000 as well?}}
	\label{fig:det_voc}
\end{figure}

\subsubsection{VOC2007 dataset}
The first experiment is conducted on the standard VOC2007 dataset \cite{Everingham10} with 20 classes. It contains 5011 images for training and validation and 4952 images for testing. We used VOC2007 trainval set to train the model and evaluate our model on the VOC2007 test set. The accuracy is measured by the standard mean average precision (mAP) at the intersection over union (IoU) of 0.5. 
%VOC2007 consists of around 5K training images with 20 categories and another 5K images as the test images. 
Results are shown in Fig~\ref{fig:det_voc}, RGC model initialization shows significant gains in both object detection algorithms as well as faster convergence speed. 

With YOLO V2, the RGC initialized model achieves mAP of $41\%$ after $50$ iterations of training, and mAP of $67\%$ after $1,000$ iterations, which is 10 times faster than the baseline. 
After the same number of $10,000$ iterations, it outperforms the baseline algorithm by $5\%$ in terms of mAP. 

With Faster R-CNN, the RGC initialized model achieves mAP of $44\%$ after $100$ iterations of training. Compared with the baseline algorithm, it uses only $2,000$ iterations to achieve mAP of $60\%$, which is $\mathbf{17}$ times faster. With the same number of $35$K iterations, our scheme eventually achieves $65\%$ mAP, which is $4\%$ higher than the baseline in terms of mAP.

\begin{figure}[h]
	\centering
    \subfloat[YOLOV2]{\includegraphics[width=0.9\linewidth] 
{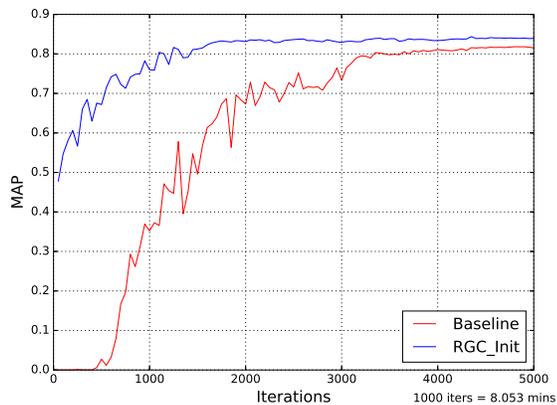}} \\    
    \subfloat[Faster R-CNN]{\includegraphics[width=0.9\linewidth]{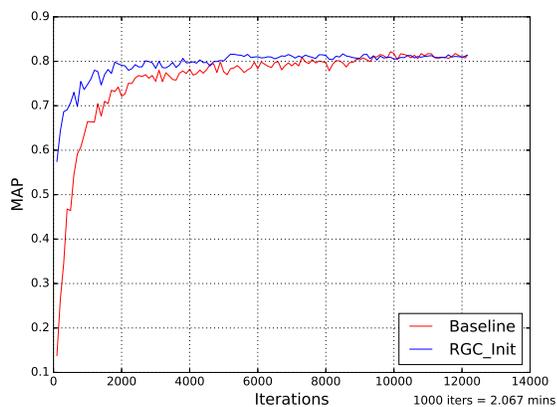}}
    \caption{Model initialization for (a) YOLO V2 and (b) Faster R-CNN on Flickr Logo32 dataset}
    \label{fig:det_flickrlogo}
\end{figure}

\subsubsection{Flickr Logo32 dataset}
The PASCAL VOC2007 dataset \cite{pascal-voc-2007} and the ImageNet dataset \cite{deng2009imagenet} are very similar in both distribution and context, we are interested to see the performance of our model initialization method when applied to another domain. We use the Flickr-Logo32 \cite{Flickrlogo} for logo detection. It contains $2240$ images of $32$ brands (we removed 6000 non-logo images). We further split the data set into a training set with 1736 images and a testing set with 434 images. 

Results are shown in {Fig~\ref{fig:det_flickrlogo} }, we observe that the RGC model initialization method consistently improves YOLO V2 with a $4$ times speedup and $2.4\%$ mAP gain in the final model. It also helps the convergence of Faster R-CNN, with a $2.4$ times speed up but marginal gain in mAP in the final model. We argue that the decrease of gain in Faster RCNN might be due to its algorithm structure. As pointed out in \cite{redmon2016you}, YOLO is better at transferring to other domains than Faster RCNN. However, the consistent improvements in convergence speed demonstrates the effectiveness of our proposed method.

\section {Conclusion}
\label{sec:conclusion}
In this paper, we have presented a regularized Gaussian classifier (RGC) with closed-form solution to initializing the last linear layer of a DNN model, which was normally randomly initialized due to the lack of analytical solution of logistic regression. This model initialization algorithm significantly reduces the DNN model fine-tuning cost and also leads to a better model, as validated on both image classification and object detection tasks. We also showed that properly initialized model parameters can reduce the model variance which is the main reason for the performance improvement. There are still some problems not addressed in this paper. For example, 1) how to initialize the model for regression tasks which is crucial for many vision tasks, such as face alignment and bounding box regression; 2) how to regularize the weights which also minimize the cross-entropy loss. These will be our future research directions.

\appendix
\section{Appendix}
\subsection{Discussion on over-fitting}

In this work, we have shown that RGC-based model initialization leads to faster convergence, which is easy to understand as it initializes a CNN model with an approximate rather than random solution. But we also found that this approach also leads to better accuracy, which is counter-intuitive as logistic regression is known to be convex and any initialization should lead to the same global optimal solution.

The understand this, we performed the following two experiments to study why RGC-initialized models are less likely to over-fit.

\begin{figure}
	\centering
	\subfloat[Weight decay 0.0005]{\includegraphics[width=0.9\linewidth]{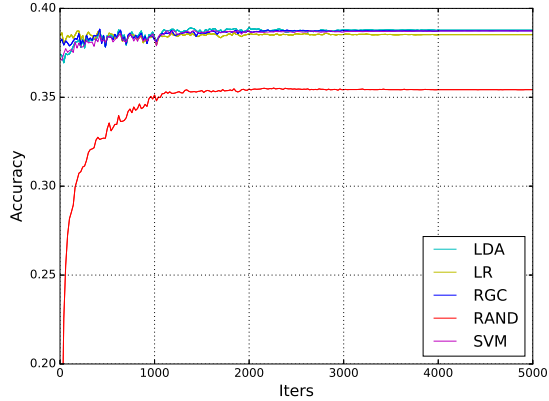}}  \\
	\subfloat[Weight decay 0.05] {\includegraphics[width=0.9\linewidth]{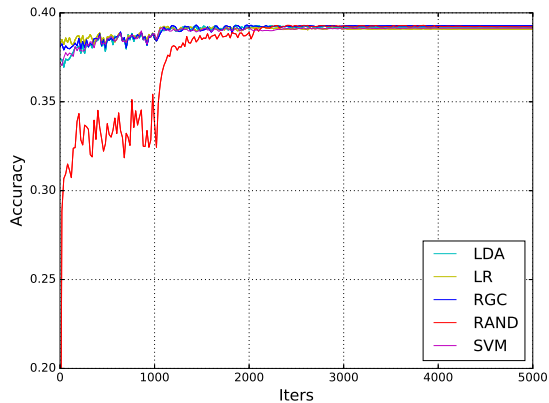}}
	\caption{Fine-tune the last layer for the Flickr-style task with different model initialization methods. }
	\label{fig:flickrstyle_fl}
\end{figure}

In the following experiments, we fix all layers other than the newly initialized one. First, we fine-tune the new layer with learning rate 0.01 and weight decay 0.0005 and fine-tune the model for 5000 iterations. Results are shown in Fig. \ref{fig:flickrstyle_fl} (a). We can see that the performance of data-dependent model initialization methods are similar which are better than the RAND initialization by nearly 3.0\%.

To demonstrate the degradation of RAND initialization is caused by over-fitting, we perform the second experiment by increasing the weight decay 100 times to impose a stronger regularization and results are shown in Fig. \ref{fig:flickrstyle_fl} (b). Increasing weight decay does not have much effect on data-dependent initialization methods but significantly improves the performance of RAND initialization. It means some of the gap in the first setting is caused by over-fitting.

These experiments show that RGC along with other data-dependent initialization methods prevent model from over-fitting.

% \leizhangc{this over-fitting study looks quite unexpected. We should add some introductory words to bring in this topic.} If we fix all the other parameters and only fine-tune the last layer (we use the same set of training parameters as \cite{jia2014caffe} with weight decay: 0.0005, step size: 1000, batch size: 256, learning rate: 0.01 to fine-tune the model for 5000 iterations), the model is likely to over-fit on the training set. 
% From Fig. \ref{fig:flickrstyle_fl}.a, we can see that the performance of different data-dependent model initialization method are quite similar. The RGC is slightly better than other algorithms by 0.1 percent. We can also find that Random-initialized method hurt the accuracy by 2.9 percent, which is remarkable. \leizhangc{please use hyphen consistently in the paper between random and initialized, and between data and dependent.}

Fine-tuning only the last layer is equivalent to solving multinomial logistic regression with an iterative solver (\emph{e.g.} SGD). Since the cross entropy of an exponential family is always convex, multinomial logistic regression is convex and has a unique global minimum. This is why all data-dependent initialization end up with similar accuracy. However, an interesting problem is why RAND initialization behaves differently?

\begin{figure}
	\centering
    \subfloat[training and testing loss over different iterations]{\includegraphics[width=0.9\linewidth]{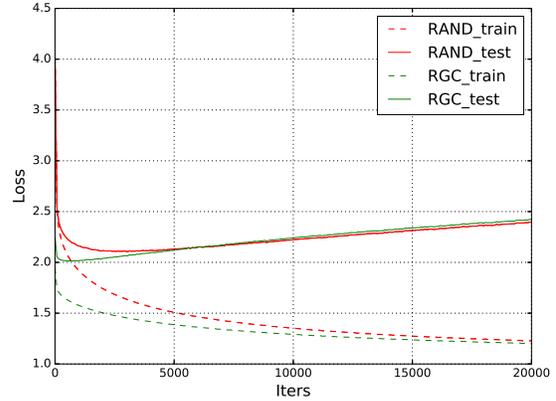}}  \\
	\subfloat[testing loss over training loss]{\includegraphics[width=0.9\linewidth]{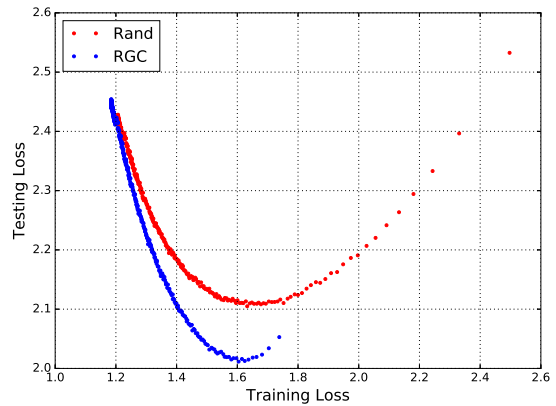}}
	\caption{Training and testing loss on Flickr-style task (weight decay=0.0005). The red curves represent the results from the randomly initialized method, the green curves represent the results from the RGC model initialization.}% \leizhangc{Fig. b has blue curves not green curves. Should we use the same color in a and b?}}
	\label{fig:overfit}
\end{figure}

To address this problem, we fix the learning rate to 1e-4 and continue training the model for 20k iterations. Results are shown in Fig. \ref{fig:overfit} (a), both training losses and testing losses of RGC-initialized model and RAND-initialized model are converged. We also find from Fig. \ref{fig:overfit} (b) that RGC-initialized model has smaller training loss to testing loss ratio which means it is less likely to over-fit than the RAND-initialized model.

\begin{figure}
	\centering
	\includegraphics[width=0.95\linewidth]{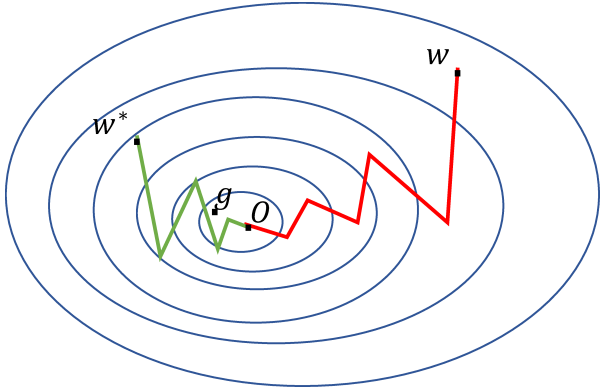}
	\caption{Better initialization method helps SGD optimization. $O$ is the minimum on the training set, $G$ is the minimum on the test set, $W$ is the randomly initialized weight, $W^*$ is the RGC-initialized weight. }
	\label{fig:sgdopt}
\end{figure}

Compared with Fig. \ref{fig:flickrstyle_fl} (a) and Fig. \ref{fig:flickrstyle_fl} (b), we find that the data-dependent model initialization methods reduce the risk of over-fitting. We observed such phenomena consistently in different experiments, which can be explained in Fig. \ref{fig:sgdopt}. SGD is a local greedy optimization algorithm. It gradually seeks for path to reduce the loss on the training set. However, due to the variance between training and testing set, SGD may be stuck early in a local minimum of the loss function defined on the training set, which is possibly not even close to the global optimal solution of the test set. A good model initialization method has the potential to move the SGD initial condition closer to the test set optima and lead to a better result. Especially when using a approximate solution, it may even place the initial condition within the same local minimum region. In contrary, RAND initialization places initial condition far from global optimal, since fine-tuning uses small learning rate, SGD is likely to stuck in a local minimum far from global optimum.

%\leizhangc{A more fundamental explanation is that RGC uses centroid statistics which has less variance between training and test sets.}

{\small
\bibliographystyle{ieee}
\bibliography{egbib}

\begin{thebibliography}{10}\itemsep=-1pt

\bibitem{cheng2018revisiting}
B.~Cheng, Y.~Wei, H.~Shi, R.~Feris, J.~Xiong, and T.~Huang.
\newblock Revisiting rcnn: On awakening the classification power of faster
  rcnn.
\newblock In {\em ECCV}, 2018.

\bibitem{dai2016r}
J.~Dai, Y.~Li, K.~He, and J.~Sun.
\newblock R-fcn: Object detection via region-based fully convolutional
  networks.
\newblock In {\em NIPS}, pages 379--387, 2016.

\bibitem{deng2009imagenet}
J.~Deng, W.~Dong, R.~Socher, L.-J. Li, K.~Li, and L.~Fei-Fei.
\newblock Imagenet: A large-scale hierarchical image database.
\newblock In {\em CVPR}, pages 248--255, 2009.

\bibitem{pascal-voc-2007}
M.~Everingham, L.~Van~Gool, C.~K.~I. Williams, J.~Winn, and A.~Zisserman.
\newblock The {PASCAL} {V}isual {O}bject {C}lasses {C}hallenge 2007 {(VOC2007)}
  {R}esults.
\newblock
  http://www.pascal-network.org/challenges/VOC/voc2007/workshop/index.html.

\bibitem{Everingham10}
M.~Everingham, L.~Van~Gool, C.~K.~I. Williams, J.~Winn, and A.~Zisserman.
\newblock The pascal visual object classes (voc) challenge.
\newblock {\em International Journal of Computer Vision}, 88(2):303--338, June
  2010.

\bibitem{rlda}
J.~H. Friedman.
\newblock Regularized discriminant analysis.
\newblock {\em Journal of the American Statistical Association},
  84(405):165--175, 1989.

\bibitem{Glorot10understandingthe}
X.~Glorot and Y.~Bengio.
\newblock Understanding the difficulty of training deep feedforward neural
  networks.
\newblock In {\em In Proceedings of the International Conference on Artificial
  Intelligence and Statistics (AISTATS’10). Society for Artificial
  Intelligence and Statistics}, 2010.

\bibitem{caffe-oxford102}
J.~Goode.
\newblock Caffe cnns for the oxford 102 flower dataset.
\newblock \url{https://github.com/jimgoo/caffe-oxford102}, 2015.

\bibitem{goodfellow6572explaining}
I.~J. Goodfellow, J.~Shlens, and C.~Szegedy.
\newblock Explaining and harnessing adversarial examples.
\newblock {\em arXiv preprint arXiv:1412.6572}.

\bibitem{griffinHolubPerona}
G.~Griffin, A.~Holub, and P.~Perona.
\newblock Caltech-256 object category dataset.
\newblock Technical Report 7694, California Institute of Technology, 2007.

\bibitem{guo2016msceleb}
Y.~Guo, L.~Zhang, Y.~Hu, X.~He, and J.~Gao.
\newblock M{S}-{C}eleb-1{M}: A dataset and benchmark for large scale face
  recognition.
\newblock In {\em ECCV}. Springer, 2016.

\bibitem{he2017mask}
K.~He, G.~Gkioxari, P.~Doll{\'a}r, and R.~Girshick.
\newblock Mask r-cnn.
\newblock In {\em ICCV}, pages 2980--2988, 2017.

\bibitem{he2015}
K.~He, X.~Zhang, S.~Ren, and J.~Sun.
\newblock Delving deep into rectifiers: Surpassing human-level performance on
  imagenet classification.
\newblock In {\em ICCV}, pages 1026--1034, 2015.

\bibitem{MSRA_150}
K.~He, X.~Zhang, S.~Ren, and J.~Sun.
\newblock Deep residual learning for image recognition.
\newblock In {\em Proceedings of the IEEE conference on computer vision and
  pattern recognition}, pages 770--778, 2016.

\bibitem{herdin2005correlation}
M.~Herdin, N.~Czink, H.~Ozcelik, and E.~Bonek.
\newblock Correlation matrix distance, a meaningful measure for evaluation of
  non-stationary mimo channels.
\newblock In {\em Vehicular Technology Conference, 2005. VTC 2005-Spring. 2005
  IEEE 61st}, volume~1, pages 136--140. IEEE, 2005.

\bibitem{jia2014caffe}
Y.~Jia, E.~Shelhamer, J.~Donahue, S.~Karayev, J.~Long, R.~Girshick,
  S.~Guadarrama, and T.~Darrell.
\newblock Caffe: Convolutional architecture for fast feature embedding.
\newblock In {\em ACM Multimedia}, pages 675--678. ACM, 2014.

\bibitem{Flickr}
S.~Karayev, M.~Trentacoste, H.~Han, A.~Agarwala, T.~Darrell, A.~Hertzmann, and
  H.~Winnemoeller.
\newblock Recognizing image style.
\newblock In {\em BMVC}, 2014.

\bibitem{krahenbuhl2015data}
P.~Kr{\"a}henb{\"u}hl, C.~Doersch, J.~Donahue, and T.~Darrell.
\newblock Data-dependent initializations of convolutional neural networks.
\newblock In {\em ICLR}, 2016.

\bibitem{AlexHinto_DNN}
A.~Krizhevsky, I.~Sutskever, and G.~E. Hinton.
\newblock Imagenet classification with deep convolutional neural networks.
\newblock In {\em NIPS}, pages 1097--1105. MIT Press, 2012.

\bibitem{lin2017feature}
T.-Y. Lin, P.~Doll{\'a}r, R.~Girshick, K.~He, B.~Hariharan, and S.~Belongie.
\newblock Feature pyramid networks for object detection.
\newblock In {\em CVPR}, volume~1, page~4, 2017.

\bibitem{liu2016ssd}
W.~Liu, D.~Anguelov, D.~Erhan, C.~Szegedy, S.~Reed, C.-Y. Fu, and A.~C. Berg.
\newblock Ssd: Single shot multibox detector.
\newblock In {\em ECCV}, pages 21--37, 2016.

\bibitem{Flower}
M.-E. Nilsback and A.~Zisserman.
\newblock Automated flower classification over a large number of classes.
\newblock In {\em Proc. of the Indian Conf. on Computer Vision, Graphics and
  Image Processing}, 2008.

\bibitem{redmon2016you}
J.~Redmon, S.~Divvala, R.~Girshick, and A.~Farhadi.
\newblock You only look once: Unified, real-time object detection.
\newblock In {\em CVPR}, pages 779--788, 2016.

\bibitem{yolo9000}
J.~Redmon and A.~Farhadi.
\newblock Yolo9000: better, faster, stronger.
\newblock In {\em CVPR}, 2017.

\bibitem{frcnn}
S.~Ren, K.~He, R.~Girshick, and J.~Sun.
\newblock Faster r-cnn: Towards real-time object detection with region proposal
  networks.
\newblock In {\em NIPS}, pages 91--99, 2015.

\bibitem{ren2015faster}
S.~Ren, K.~He, R.~Girshick, and J.~Sun.
\newblock Faster r-cnn: Towards real-time object detection with region proposal
  networks.
\newblock In {\em NIPS}, pages 91--99, 2015.

\bibitem{Flickrlogo}
S.~Romberg, L.~G. Pueyo, R.~Lienhart, and R.~van Zwol.
\newblock Scalable logo recognition in real-world images.
\newblock In {\em Proc. of the 1st ACM Int'l Conf. on Multimedia Retrieval},
  pages 25:1--25:8, New York, NY, USA, 2011.

\bibitem{ILSVRC15}
O.~Russakovsky, J.~Deng, H.~Su, J.~Krause, S.~Satheesh, S.~Ma, Z.~Huang,
  A.~Karpathy, A.~Khosla, M.~Bernstein, A.~C. Berg, and L.~Fei-Fei.
\newblock {ImageNet Large Scale Visual Recognition Challenge}.
\newblock {\em International Journal of Computer Vision}, 115(3):211--252,
  2015.

\bibitem{seuret2017pca}
M.~Seuret, M.~Alberti, M.~Liwicki, and R.~Ingold.
\newblock Pca-initialized deep neural networks applied to document image
  analysis.
\newblock In {\em Document Analysis and Recognition (ICDAR), 2017 14th IAPR
  International Conference on}, volume~1, pages 877--882. IEEE, 2017.

\bibitem{vgg_DNN}
K.~Simonyan and A.~Zisserman.
\newblock Very deep convolutional networks for large-scale image recognition.
\newblock {\em arXiv preprint abXiv:1409.1556}, 2014.

\bibitem{randominit}
I.~Sutskever, J.~Martens, G.~Dahl, and G.~Hinton.
\newblock On the importance of initialization and momentum in deep learning.
\newblock In {\em ICML}, pages 1139--1147, 2013.

\bibitem{tibshirani2002diagnosis}
R.~Tibshirani, T.~Hastie, B.~Narasimhan, and G.~Chu.
\newblock Diagnosis of multiple cancer types by shrunken centroids of gene
  expression.
\newblock {\em Proceedings of the National Academy of Sciences},
  99(10):6567--6572, 2002.

\bibitem{wei2018ts2c}
Y.~Wei, Z.~Shen, B.~Cheng, H.~Shi, J.~Xiong, J.~Feng, and T.~Huang.
\newblock Ts2c: Tight box mining with surrounding segmentation context for
  weakly supervised object detection.
\newblock In {\em European Conference on Computer Vision}, 2018.

\bibitem{Yoshua:nips2014}
J.~Yosinski, J.~Clune, Y.~Bengio, and H.~Lipson.
\newblock How transferable are features in deep neural networks?
\newblock In {\em NIPS}, pages 3320--3328, 2014.

\bibitem{zeiler2014visualizing}
M.~D. Zeiler and R.~Fergus.
\newblock Visualizing and understanding convolutional networks.
\newblock In {\em European conference on computer vision}, pages 818--833.
  Springer, 2014.

\bibitem{Places}
B.~Zhou, A.~Lapedriza, J.~Xiao, A.~Torralba, and A.~Oliva.
\newblock Learning deep features for scene recognition using places database.
\newblock In {\em NIPS}, pages 487--495, Cambridge, MA, USA, 2014.

\end{thebibliography}
}

\end{document}